\documentclass{article}

\usepackage{microtype}
\usepackage{graphicx}
\usepackage{subcaption}
\usepackage{booktabs} 

\usepackage{hyperref}
\usepackage{xurl}
\usepackage{xcolor}

\usepackage[preprint]{icml2026}

\usepackage{amsmath}
\usepackage{amssymb}
\usepackage{mathtools}
\usepackage{amsthm}

\usepackage[capitalize,noabbrev]{cleveref}

\theoremstyle{plain}

\theoremstyle{definition}

\theoremstyle{remark}

\usepackage[textsize=tiny]{todonotes}

\icmltitlerunning{``Sorry, I Didn't Catch That'': How Speech Models Miss What Matters Most}

\begin{document}

\twocolumn[
  \icmltitle{\textit{``Sorry, I Didn't Catch That''}:\\ How Speech Models Miss What Matters Most}

  \begin{icmlauthorlist}
    \icmlauthor{Kaitlyn Zhou}{1,2}
    \icmlauthor{Martijn Bartelds}{1,3}
    \icmlauthor{Federico Bianchi}{1}
    \icmlauthor{James Zou}{1,3}
  \end{icmlauthorlist}

  \icmlaffiliation{1}{TogetherAI}
  \icmlaffiliation{2}{Cornell University}
  \icmlaffiliation{3}{Stanford University}

  \icmlcorrespondingauthor{Kaitlyn Zhou}{kzhou@together.ai}

  \icmlkeywords{Machine Learning, Speech, Dataset}

  \vskip 0.3in
]

\printAffiliationsAndNotice{}  

\begin{abstract}
Despite speech recognition systems achieving low word error rates on standard benchmarks, they often fail on short, high-stakes utterances in real-world deployments. Here, we study this failure mode in a high-stakes task: the transcription of U.S. street names as spoken by U.S. participants. We evaluate 15 models from OpenAI, Deepgram, Google, and Microsoft on recordings from linguistically diverse U.S. speakers and find an average transcription error rate of 44\%. We quantify the downstream impact of failed transcriptions by geographic locations and show that mis-transcriptions systematically cause errors for all speakers, but that routing distance errors are \textbf{twice} as large for non-English primary speakers compared to English primary speakers. To mitigate this harm, we introduce a synthetic data generation approach that produces diverse pronunciations of named entities using open-source text-to-speech models. Fine-tuning with less than 1000 synthetic samples improves street name transcription accuracy by nearly 60\% (relative to base models) for non-English primary speakers. Our results highlight a critical gap between benchmark performance and real-world reliability in speech systems and demonstrate a simple, scalable path to reducing high-stakes transcription errors. 
\end{abstract}

\section{Introduction}
Speech recognition systems are deployed in applications where transcription errors have immediate consequences, from ride-hailing to emergency dispatch \cite{twilio2025curb, cisa2025ai}. In this paper, we examine a key limitation of modern systems: the frequent failure to accurately transcribe street names. A street name anchors a request to a precise physical location and is often the primary piece of information used to route responders, dispatch drivers, or deliver assistance. Because even small transcription errors can misdirect resources or delay help, we focus on a simple but high-stakes question: when a U.S. speaker provides a street name by voice, can a deployed system reliably transcribe it? We evaluate 15 models from the top-performing speech recognition providers—OpenAI, Deepgram, Google, and Microsoft and find that 44\% of the street names are erroneous. In other words, almost every other street name given to these models will be incorrectly transcribed.

\begin{figure}[t]
    \centering
    \includegraphics[width=\linewidth]{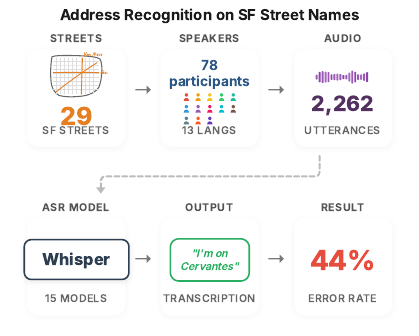}
    \caption{Overview of Transcription Evaluation Pipeline}
    \label{fig:placeholder}
\end{figure}

To illustrate the real-world implications of these transcription errors, consider the case of taxi services, where accurate street name recognition directly determines pickup and drop-off locations. Our findings reveal that street name transcription accuracy differs greatly across demographic groups and is consistently lower for speakers whose primary language is not only English.\footnote{Participants can have multiple primary languages, such as English and Spanish, which would be considered "not only English" primary language, \S \ref{sec:recruitment}} To quantify the practical and potentially financial impact of these disparities in a ride-hailing setting, we query a map API using the generated transcriptions and measure the resulting distance from the intended destination. We find that the distance between the generated and intended locations is nearly \textbf{two} times larger for non–English primary speakers than for English-only primary speakers. Although taxi usage is uneven across demographic groups \cite{jiang2019ridehailing, sikder2019uses}, taxis play a vital role for elderly, low-income, and disabled populations by providing an often subsidized means of transportation \cite{kaufman2016intelligent}. If speech-to-text systems were used in San Francisco for ride-hailing, we estimate that the average passenger pickup would be misrouted by 1.26 miles, equivalent to roughly 5 minutes of city driving time to correct the error (without additional overhead). When we scale this per-error delay by the city’s annual taxi volume of voice-based street name entries and our measured error rates, we estimate roughly \textbf{43,500} hours of avoidable delay per year, about \textbf{5} years of continuous waiting time.

Motivated by these findings, we introduce a new training recipe to generate synthetic speech with varied pronunciations of named entities, making it possible to finetune models on any named entity (e.g., street names, hospitals, descriptive locations). Our pipeline takes advantage of open-sourced text-to-speech models (specifically \texttt{Coqui-TTS}) and a public dataset (\texttt{Common Voice}) --- making it accessible for future practitioners to re-implement this strategy. With less than 1000 synthetic speech samples, we observe an improvement in street name error of nearly 60\% (relative to base error) among speakers who are not English-only primary speakers. 

We release both datasets as benchmarks for the community: SF Streets, containing 2,262 utterances from 78 participants, and US Streets, containing 3,600 recordings of 360 street names across 12 major U.S. cities from 97 unique participants. In sum, we make the following contributions:
\vspace{0em}
\begin{itemize}
    \itemsep-0.2em 
    \item We uncover a weakness in state-of-the-art speech systems: the inability to accurately transcribe information that directly affects resource allocation.
    \item We show that transcription error rates differ substantially across speakers, with notably lower accuracy for non-English-only primary speakers.
    \item We introduce a practical, open-source approach for generating synthetic speech data, achieving nearly 60\% improvement in named entity transcription performance.
    \item We curate and release two datasets, one comprised of 78 unique speakers with 2,262 of S.F. streets and then another of 97 speakers with 3,600 utterances of street names from 12 major U.S. cities.\footnote{Code: \url{https://github.com/kzhou-cloud/sf_streets_public} Dataset: \url{https://huggingface.co/datasets/kzhou/sf_streets}}
\end{itemize}

\section{Background and Related Work}

Recent advances in multi-modal models have made speech recognition systems a routine part of everyday life. One sector that has embraced the cost-saving capabilities of speech models is in live agent call centers. Speech models are ubiquitously deployed in lieu of live agents from ride-hailing companies and emergency call centers. For example, Curb---the most popular taxi application in the U.S.---leverages speech models from DeepGram and Google, resulting in an 80\% reduction in call transfers to live agents \cite{twilio2025curb, twilio_gather}. Similarly, beginning June 2024, the Metro Nashville Department of Emergency Communications uses a real-time speech-to-text transcription to handle calls \cite{cisa2025ai}. Although there are clear economic incentives at play to introduce these systems, the actual safety risks and allocation biases of the systems have yet to be systematically studied. However, in real-world deployment settings such as ride-hailing or emergency call centers --- how well are these speech models actually doing?

Speech recognition systems have long been evaluated on standard benchmarks such as Switchboard \cite{godfrey1992switchboard}, WSJ \cite{paul1992design}, TIMIT \cite{garofolo1993timit}, CALLHOME \cite{canavan1997callhome}, Fisher \cite{cieri2004fisher}, Librispeech \cite{panayotov2015librispeech} --- with many of these benchmarks having single-digit \textit{word error rate} (WER) among state-of-the-art models. We've seen tremendous progress in automatic speech recognition (ASR) systems as models continue to scale. More recently, the speech field has started to focus on the challenge of named entity recognition, such as developing a named entity corrector \cite{garg2020hierarchical}, developing end-to-end NER extraction \cite{ghannay2018end}, multilingual NER \cite{ning2024breaking}, among other survey works \cite{caubriere2020we}. This area of research remains nascent, and most state-of-the-art leaderboards have yet to incorporate these new, challenging tasks.
\section{Dataset}
In our work, we seek to augment existing automatic speech recognition (ASR) and contribute to the emerging area of named entity recognition in speech models. In particular, we evaluate the performance of deployed state-of-the-art speech models on their ability to recognize real U.S. street names, as spoken by U.S.-based participants. In order to perform this evaluation, we contribute two datasets:

\paragraph{SF Streets Dataset} Our first dataset consists of recordings from U.S.-based participants pronouncing San Francisco street names.

\begin{figure}[h!]
    \centering
    \includegraphics[width=0.95\linewidth]{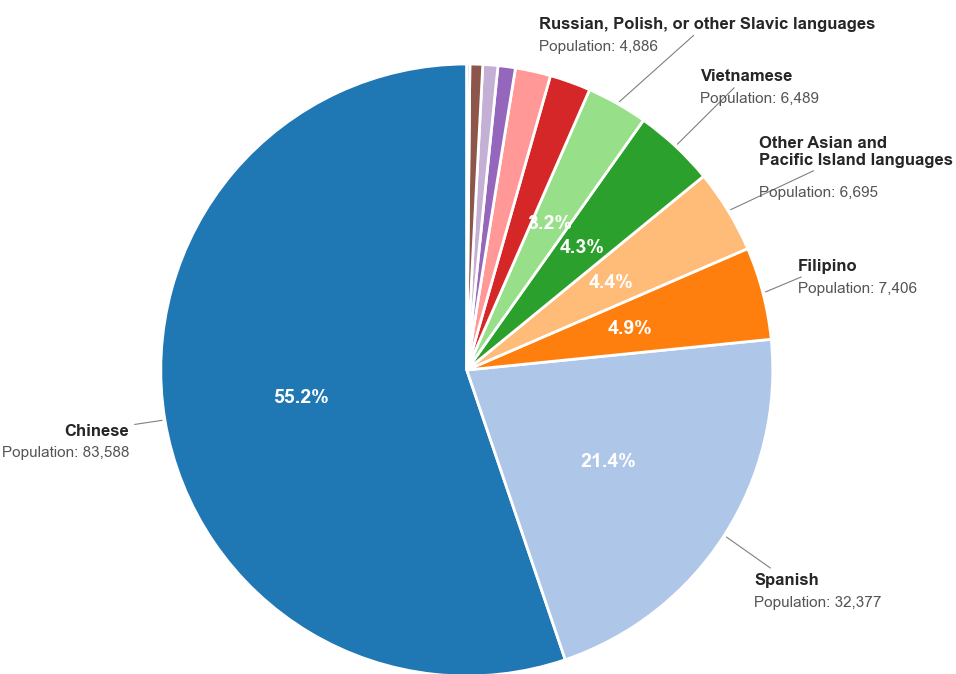}
    \caption{Limited English Proficiency Speakers in San Francisco. Original data from \cite{SF_LanguageDiversityData_SFgov}.}
    \label{fig:lep_sf}
\end{figure}

San Francisco is part of a metropolitan area of approximately 4.5 million residents and has the fifth-highest GDP among U.S. metropolitan regions \cite{us_census}. The SF street names were selected as the city also has a large population of non-native English speakers (\citeauthor{sfgov_language_diversity}. Figure \ref{fig:lep_sf}) and a well-documented historical record of street-name origins \cite{carlisle_sf_streets}. We use all boulevards of San Francisco ($n=29$) (e.g., Cesar Chavez, Alemany), which are widely recognized, frequently referenced, and likely to appear in spoken navigation and location queries.\footnote{All unique boulevard names were included, except \textit{``Skyline Boulevard''} which was accidentally omitted.} Unless otherwise noted, all evaluations and experiments of this paper are based on the SF Streets Dataset. We release a public version of this dataset as a benchmark for the community, details in \S \ref{sec:sf_streets_public}.

\paragraph{Participants}
\label{sec:recruitment}
Participants were recruited via Prolific ($n=78$) to obtain a linguistically diverse sample that reflects San Francisco’s population, with IRB approval. We collected basic demographic information, including age, gender, and race, as well as participants’ primary spoken language (\textit{Which of the following is your primary spoken language?''}), which can include multiple languages. Participants are grouped into three categories based on their response: \textcolor[HTML]{599ebc}{English Only}, indicating English as the sole primary language; \textcolor[HTML]{e68600}{Multilingual w/ English}, indicating multiple primary languages including English (e.g., English and Spanish); and \textcolor[HTML]{b4638d}{Non-English}, indicating primary language(s) other than English (e.g., Spanish and Portuguese).

A total of 80 participants were initially recruited; 2 were excluded due to very low-quality audio recordings caused by background noise or static, as determined by manual review (Table \ref{tab:demographics}). Our manual evaluation also helped verify that the voice recordings discernibly pronounced the street names appropriately and that errors from incorrect pronunciations were unlikely. Notably, all participants spoke and read English and were able to complete the recording tasks. 

For each street name, participants recorded a single utterance using the template phrase \textit{``I'm on [STREET NAME]''}. The study took approximately 8 minutes to complete, and participants were paid $\$15$ USD an hour. A total of 2,262 ($78 * 29$) utterances were collected. 

We evaluated fifteen models on this dataset:  OpenAI's {Whisper tiny (39M), base (74M), small (244M), medium (769M), large (1.5B)}, Deepgram's {nova-2}*, {nova-3}*, {telephony}*, {base-phonecall}*, {base-general}*, {enhanced-general}*, {enhanced-phonecall}*, Microsoft's {phi-4-multimodal (14B)},  and Google's {Chirp 2 (2B)}, {Chirp 3}*.\footnote{*unknown model size} Unless otherwise noted, all figures and results in the main text are computed on this dataset; evaluations on our second dataset are reported in the Appendix. In addition to being state-of-the-art models, many of these models are deployed in the real world and made for production and enterprise-scale speech recognition. Deepgram’s Nova has versions of telephony and phonecall-optimized variants, which are explicitly tuned for real-world domains (e.g., low-bandwidth telephony audio), making them strong comparison baselines.

\begin{table}[t]
\centering
\small
\renewcommand{\arraystretch}{1.1}
\begin{tabular}{@{}llrr@{}}
\toprule
\textbf{SF Streets} &  & $n$ & \% \\
\midrule
\textit{Gender} & Male & 37 & 48 \\
 & Female & 40 & 51 \\
 & Prefer not to say & 1 & 1 \\
\addlinespace
\textit{Age} & 20--29 & 19 & 24 \\
\textit{(M=38.1, SD=11)} & 30--39 & 29 & 37 \\
 & 40--49 & 17 & 22 \\
 & 50--59 & 9 & 12 \\
 & 60--69 & 3 & 4 \\
 & 70+ & 1 & 1 \\
\addlinespace
\textit{Primary Language} & \textcolor[HTML]{599ebc}{English Only} & 19 & 24 \\
 & \textcolor[HTML]{e68600}{Multilingual w/ English} & 49 & 63 \\
 & \textcolor[HTML]{b4638d}{Non-English} & 10 & 13 \\
\bottomrule
\end{tabular}
\caption{Participant demographics for SF streets dataset ($n=78$). The participants' primary languages represented 13 unique languages (Vietnamese, French, Spanish, Polish, English, Arabic, Portuguese, Korean, Chinese, Tagalog/Filipino, Russian, Japanese, and German).}
\label{tab:demographics}
\end{table}

\paragraph{U.S. Streets Dataset} The second dataset is an extension of the first dataset, which contains 30 randomly selected non-numerical street names for 12 major and diverse U.S. cities. For added variation, the street names are preceded by one of 18 prefixes, slightly increasing the difficulty of the dataset, Table \ref{tab:paraphrases}. Participants were recruited from Prolific, and all were non-English primary speakers. The final dataset contains 3,600 voice recordings from 97 participants, with each street pronounced 10 times. The study took approximately 10 minutes to complete, and participants were paid \$18 USD an hour. The average accuracy of this dataset among Whisper models of all sizes is around 24\%. We release this dataset for public research use. 

\begin{table}[h!]
\centering
\small
\renewcommand{\arraystretch}{1.1}
\begin{tabular}{@{}llrr@{}}
\toprule
\textbf{US Streets} &  & $n$ & \% \\
\midrule
\textit{Gender} & Male & 52 & 54 \\
 & Female & 45 & 46 \\
 & Prefer not to say & 0 & 0 \\
\addlinespace
\textit{Age} & 20--29 & 32 & 33 \\
\textit{(M=37.2, SD=11.3)} & 30--39 & 29 & 30 \\
 & 40--49 & 21 & 22 \\
 & 50--59 & 11 & 11 \\
 & 60--69 & 4 & 4 \\
 & 70+ & 0 & 0 \\
\addlinespace
\textit{Primary Language} & \textcolor[HTML]{599ebc}{English Only} & 0 & 0 \\
 & \textcolor[HTML]{e68600}{Multilingual w/ English} & 72 & 74 \\
 & \textcolor[HTML]{b4638d}{Non-English} & 25 & 26 \\
\bottomrule
\end{tabular}
\caption{Participant demographics for U.S. Streets Dataset ($n=97$). The participants' primary languages represented 29 unique languages (Chinese, Farsi, English, Italian, Greek, Romanian, Korean, Punjabi, Hakka, Arabic, Faroese, Croatian, Tamil, Urdu, Indonesian, German, Malayalam, Gujarati, Hindi, Spanish, Laotian, Russian, Japanese, Polish, Thai, French, Other, Portuguese, Vietnamese).}
\label{tab:demographics}
\end{table}

\paragraph{Metrics: Transcription Error Rate}
Word Error Rate (WER) is a widely used metric in speech recognition that measures the overlap between a ground-truth transcript and a model’s output. However, our findings show that WER provides an incomplete picture of transcription quality, particularly for critical information. Here, we calculate \textit{transcription error rate}, the rate at which the transcribed street name phonetically matches the target, meaning orthographically agnostic (e.g., both ``Ceasar'' and ``Cesar'' are considered correct). For SF Streets dataset, this was done by manually going through all the transcriptions produced by models and selecting phonetic equivalents. For transparency, we include this short list ($n=12$) of aliases in Table \ref{tab:SF_streets_alternatives}.

\section{Street Name Recognition Is Challenging}
\begin{figure}[t]
    \centering
    \includegraphics[width=0.95\linewidth]{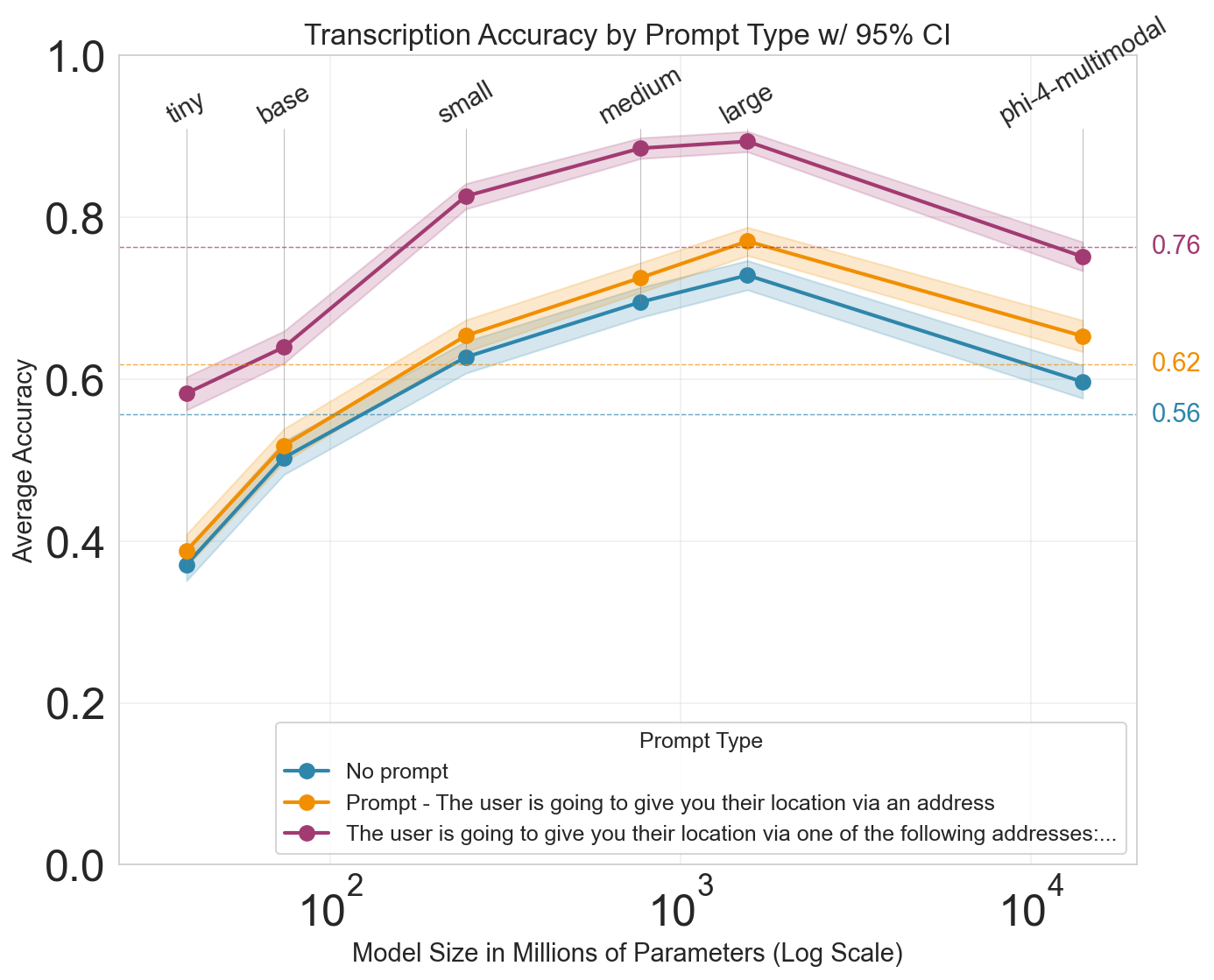}
    \caption{Overall Transcription Accuracy on SF Streets for Models That Accept a Prompt}
    \label{fig:finding_1}
\end{figure}

Our first finding is that street name recognition continues to be a major failure mode even for relatively large speech models. Across model families and prompting strategies (Figure \ref{fig:finding_1}), we find that models at every size make frequent mistakes on named entities. On the SF Streets dataset \texttt{Whisper-Base} achieves only a little over 50\% accuracy on average. Although accuracy improves with scale, the gains come with steep deployment costs. For instance, \texttt{Whisper-Large} (1.5B) reaches roughly 73\% accuracy but runs about 7× slower than \texttt{Whisper-Base} and consumes around 10× more virtual memory, substantially complicating real-world use. These high error rates also contrast with our prior beliefs about these state-of-the-art systems with very low WER. In our analysis, we find that models can have low WER but still highly consequential transcription errors (e.g., \textit{``Font''} being transcribed as \textit{``Bont''} is low in edit distance, but potentially high in geographic location). For instance, Whisper-Large achieves a low overall WER of 14\%, yet its street name transcription error rate rises to 27\% when transcribing street names correctly. This gap underscores the difficulty models face with named entities and reveals the limitations of relying on standard aggregate metrics. \cite{radford2022robustspeechrecognitionlargescale}.  

\begin{figure*}[h!]
    \centering
    \includegraphics[width=0.95\linewidth]{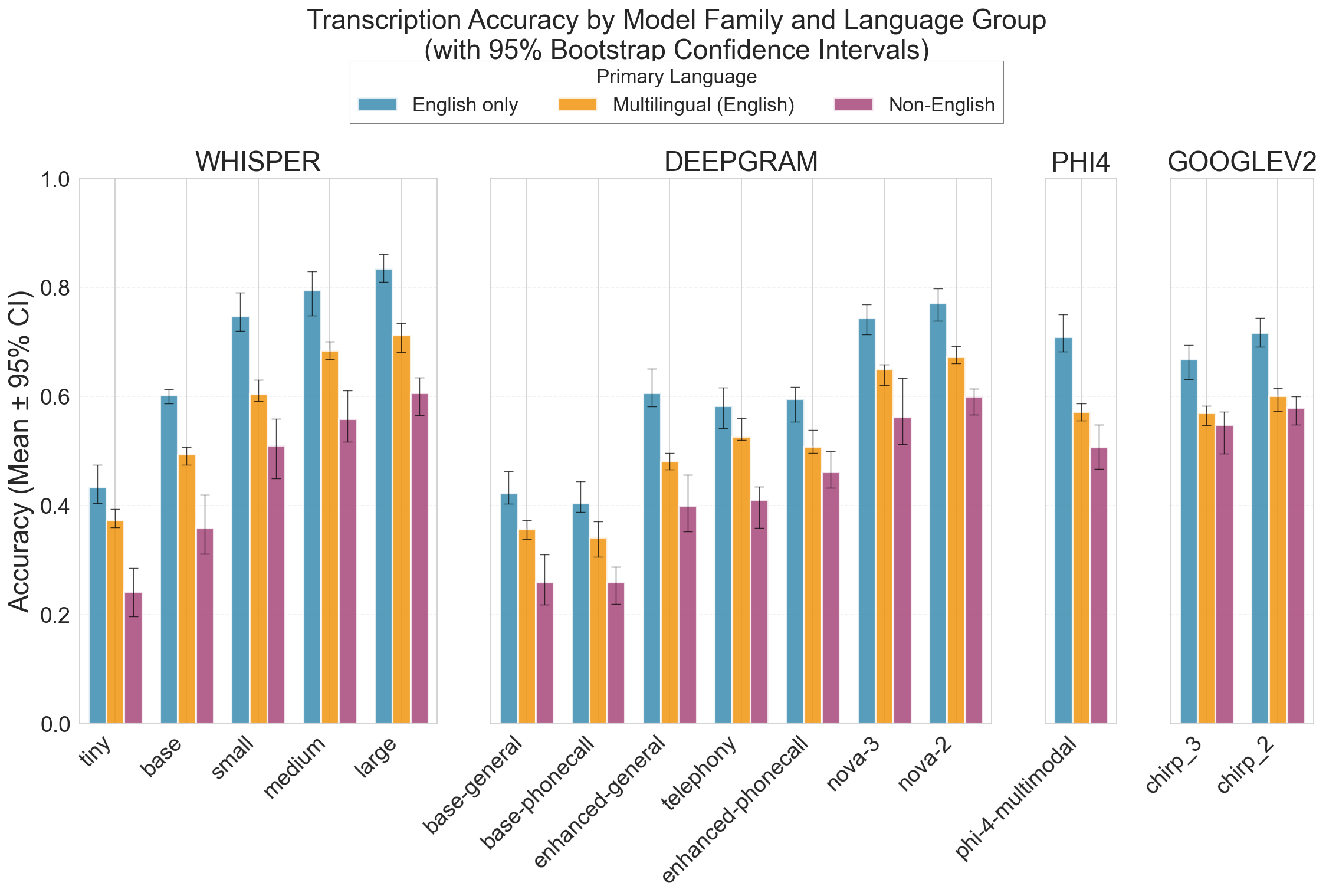}
    \caption{Transcription Accuracy by Language Groups Across All Model Families. 95\% confidence intervals calculated via bootstrap resampling of 10,000 samples}
    \label{fig:finding2}
\end{figure*}

\subsection{Adding Context} One possibility why this task is so hard is that street names may be too infrequent in training data for the model to generate them without additional context. To test this, we re-evaluated promptable models like (\texttt{Whisper} and \texttt{Phi-4}) using prompts that provide additional contextual cues. We used the following lightweight prompt in this experiment: \texttt{The user is going to give you their location via an address.} This prompt adds the kind of situational framing a real system could plausibly provide. Unfortunately, we see almost no gains from this situational awareness; although the model is more likely to guess a street name, it continues to fail to correctly identify the one that is being spoken. We then we ran a second condition that explicitly supplies the full set of target street names ($n=29$) in the system prompt: \texttt{The user is going to give you their location via one of the following addresses: Alemany, Arguello...}. This prompt is not meant to model a realistic deployment scenario, but rather serves as a diagnostic upper bound. By giving the model the relevant vocabulary upfront, we largely eliminate lack of context as an explanation, leaving recognition and selection errors (mishearing, phonetic ambiguity, and choosing the wrong candidate) as the primary remaining failure modes. Even in this ``perfect context'' setting, average accuracy across tested models is 76\%, indicating that the bottleneck is not just context, but the transcription and discrimination of similar-sounding entities.

\subsection{Implications for Speech Model Evaluation}
Our results and new dataset highlight that street name recognition remains a distinct and difficult task, and one that should be evaluated explicitly before deploying speech models in mission-critical workflows.

We see two main hypotheses for why word error rates can substantially overstate the transcription reliability of named entities. The first is that the current evaluation focuses heavily on longer-form speech, where language models can take advantage of context to fill in highly probable words. The second is that street names often have historical and foreign origins, making them less frequent in training data and more variable to pronunciation differences (e.g., an English speaker pronouncing \textit{Cesar Chavez} or \textit{Arguello}). A small analysis of street names from  U.S. cities ($n=177,155$) found that 33\% of the street names came from non-English origins.\footnote{GPT 4.1 was used to classify the origin of street names.}

Our findings illustrate that even when the overall word error rate (WER) is low, models may still fail disproportionately on the named entities that carry the most operational importance.

\section{Exacerbated Errors for Non-English Primary Speakers}
As modern systems are deployed in diverse urban environments, users may vary greatly demographically in age, gender, and linguistic background. For example, the same street name can be pronounced in substantially different ways, particularly in cities such as San Francisco, which has a large population of residents for whom English is a second language. This diversity raises a concern about the fairness of speech-to-text models: Are transcription error rates the same across demographic groups? Given the well-documented challenges of automatic speech recognition for accented speech \cite{koenecke2020racial, hofmann2024ai}, we investigate whether similar disparities arise in the context of street name recognition.

\begin{figure}[h!]
    \centering
    \includegraphics[width=0.95\linewidth]{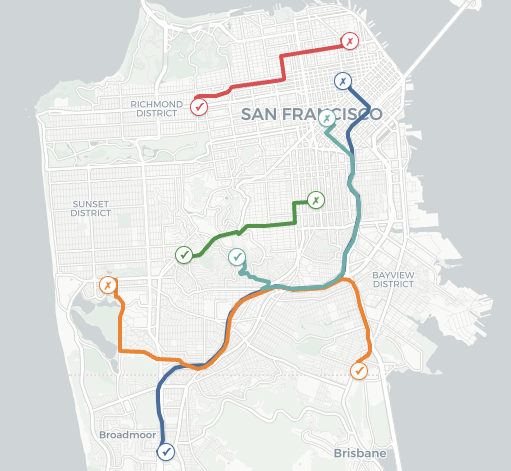}
    \caption{Visualization of the Five Worst Mistakes (by distance) of a Non-English Speaker}
    \label{fig:worst_mistakes}
\end{figure}

\subsection{Finding}
Our second key finding shows that street name recognition is not only inherently difficult, but that model errors disproportionately impact users whose primary language is not \textit{only} English. Across our 15 models and model variants, non-English primary speakers exhibited an 18\% lower accuracy compared to English primary speakers (46\% versus 64\%). Figure \ref{fig:finding2} illustrates the systemic pattern of this discrepancy. Across all four model families, we see a significant drop in accuracy between English-only primary speakers and non-English primary speakers. We observed no significant effects of self-identified gender and age on transcription performance.

\begin{figure*}[t]
    \centering
    \includegraphics[width=\linewidth]{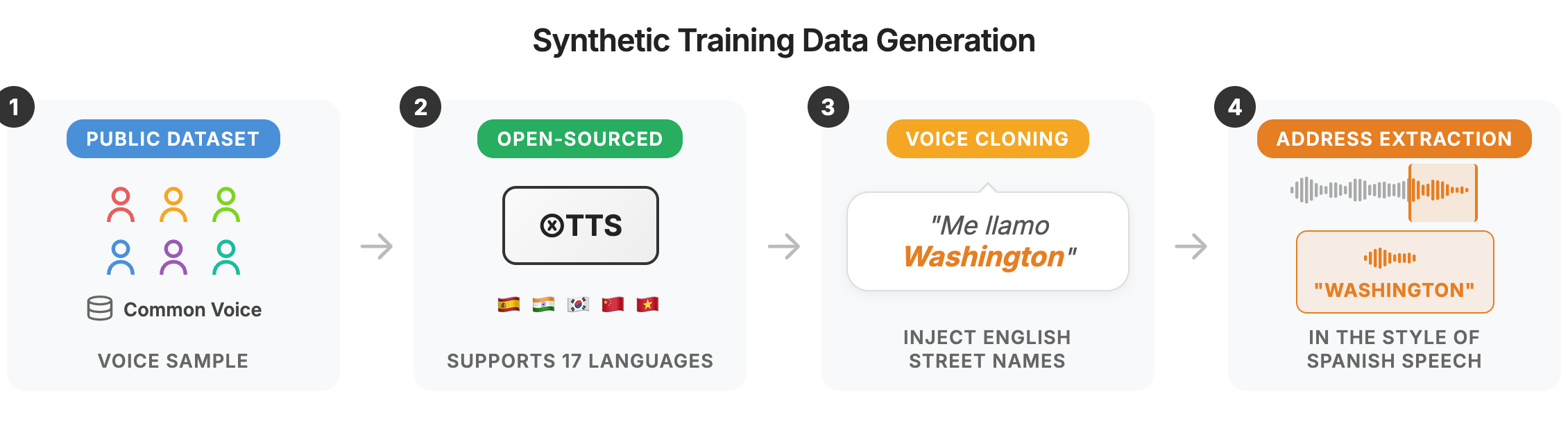}
    \caption{\textbf{(1)} Select a sample of speech from Common Voice, e.g., Spanish \textbf{(2)} Set the XTTS to generate speech in Spanish (supports 16 languages, excluding English) \textbf{(3)} Clone the voice and generate Spanish but with injected English street names, e.g., \textit{``Mi nombre es... Washington''} \textbf{(4)} Extract street name speech and manually validate. Repeat this with as many samples as needed to create a unique finetuning dataset.\footnote{Injection includes the street name and omits the ``I'm on'' prefix to preserve voice generation quality.}}
    \label{fig:workflow_2}
\end{figure*}

\subsection{Estimating the Financial Impact}
To quantify the downstream financial consequences of transcription errors, we analyze outputs from \texttt{Whisper-Base} in both English primary and non-English primary speakers by querying the Google Maps API to estimate the geographic distance between intended and transcribed destinations. Given a transcribed street name, the API returns corresponding coordinates—for example, querying ``Alemany Blvd, San Francisco'' yields the coordinates 37.72, -122.44. This approach incorporates a realistic degree of tolerance, as the API can automatically correct certain transcription errors (e.g., ``Alemony'' may still resolve to the correct location), mirroring the behavior of real-world dispatch systems. Our evaluation has some built-in leniency, giving us a conservative estimate of transcription error. First, drop instances where no location could be found even with the Google Maps API ($n=212$, 9\% of the dataset). We cap the distance error to 20 miles and discard any instances beyond this threshold ($n=6$); making the assumption that out-of-city destinations would be corrected for by humans in the loop.

Despite this added leniency, transcription errors still result in substantial geographic deviations (Figure \ref{fig:worst_mistakes}). On average, the distance between the intended and transcribed locations is 1.26 miles (in driving distance) for English primary speakers, compared to 2.4 miles for non-English primary speakers—nearly a twofold increase. Using average San Francisco taxi fares (\$0.65 per one-fifth mile; \citealp{sfmta_taxi_fares}) and an average traffic speed of 14 mph \cite{sfexaminer_traffic_2025}, these discrepancies translate into an estimated \$4.00 cost and an average expected 5-minute delay per trip for English primary speakers, versus \$8.00 and an average 10-minute expected delay for non-English primary speakers.

As of December 2025, over 6,000 taxi rides happen on a weekday basis in San Francisco \cite{sfmta_taxi_trips}. Taxis remain a critical, government-subsidized transportation service, particularly for elderly individuals and people with physical disabilities \cite{sfmta_etc}. Conservatively assuming that only one-third of weekday taxi trips involve phone-based dispatch, this gives us approximately 2,000 voice-mediated pickup requests per weekday.\footnote{While not all trips involve voice-based street name entry, survey evidence suggests that phone-based ordering remains common among frequent taxi users: a 2013 SFMTA report found that 36\% of frequent riders place the majority of their trips via phone calls, and 17\% rely on phone calls exclusively \cite{sfmta_taxi_survey}.} Using the empirically measured average delay associated with transcription errors (approximately 5 minutes per trip), this implies on the order of 43,500 hours of cumulative delay annually (2,000 trips × 261 weekdays × 5 minutes). Valued using standard taxi fare schedules, this corresponds to an estimated \$2.1 million in annual economic cost. Importantly, these estimates assume that all riders are English-primary speakers and therefore represent a lower bound on the true cost; accounting for higher observed error rates among non-English primary speakers would further increase the expected impact.
\section{Mitigation via Synthetic Data}
Motivated by the demonstrated risks and downstream consequences of transcription errors, we set out to develop a method to improve the transcription of named entities, especially for non-English primary speakers. Here, we illustrate that with existing resources, we can finetune a speech recognition model to increase performance on street name recognition using synthetic data alone. 

Although large volumes of speech data are available, it remains difficult to obtain datasets that adequately capture the wide range of pronunciation patterns exhibited by speakers with diverse language backgrounds.  Recent advances in text-to-speech (TTS) modeling present an opportunity to address this data gap and synthetically produce a broad spectrum of street name pronunciations. While such synthetic data is unlikely to fully reflect the nuances of real human speech, it may nevertheless introduce sufficient phonetic variation to improve downstream recognition performance through finetuning.

To support reproducibility and extensibility, we construct a fully reproducible training pipeline using only open-source datasets and models, enabling practitioners to adapt our approach to other cities and contexts. Specifically, we leverage the widely used XTTS model available on Hugging Face \cite{xtts} and Common Voice Scripted Speech 24.0 \cite{ardila2020common}, a multilingual speech corpus comprising 134 languages and recordings from over 350,000 speakers.

\subsection{Failed Initial Attempt}
We first tried to use XTTS to see if it was possible to clone accented speech and generate additional speech with the same pronounciation patterns. Our initial attempts were unsuccessful as XTTS generated voice clones with an American-English or British-English accent. For example, when cloning a German speaker speaking English, the synthesized output retained the speaker’s vocal characteristics but replaced the original German accent with an American-English accent. This suggests that current voice cloning techniques may systematically normalize or suppress foreign accents --- a systematic study here would be pertinent for future work.

\paragraph{Intuition and Breakthrough}
The key insight behind our synthetic data generation approach is to exploit the implicit accent style transfer of cloning models. When generating speech in a given language, the model tends to impose a canonical or ``default'' speaking style associated with that language. In our qualitative analysis, we observed that English generations were rendered with a stereotypical American-English accent, while Italian generations similarly exhibited a distinctly Italian speaking style.

Our key breakthrough was to leverage this behavior by prompting the model to generate speech in a non-English language while selectively inserting English words into the prompt. For instance, we instructed the model to generate Italian speech for the text \textit{"Buongiorno, mi chiamo... Washington"} and then isolated the audio corresponding to the English word ``Washington.'' This extraction process was automated and subsequently verified by the authors. 

\subsection{Recipe}
The final data-generation procedure for a single training example is illustrated in Figure \ref{fig:workflow_2}. The core idea is to have a voice cloning system synthesize speech in a foreign language while embedding English street names within the text, thereby inducing a form of style transfer onto the English terms.

This approach produces substantial variation in pronunciation, driven by phonetic patterns common in other languages. Additional diversity is introduced by having unique speakers per language for cloning, allowing us to perturb speaker characteristics and further expand the range of synthesized pronunciations for each street name.\footnote{Note that this approach is not mimicking stereotypical foreign accents --- rather, it relies on monolingual foreign-language speech—without English influence—and applies its phonetic structure as a style transfer onto English words.}

XTTS supports voice cloning in 16 different languages. For each of these languages, we select a speaker speaking in that language from \texttt{Common Voice Scripted Speech 24.0} (e.g., select two speakers from the Spanish dataset) and clone these voices to synthesize new text that will contain the street name (e.g., Estoy en Washington). We then automatically extract the audio that is associated with the word ``Washington'' and manually verify each of these files. We finetune \texttt{Whisper-base} (batch size 16, learning rate 1e-5, early stopping loss threshold at 0.01). With less than 1,000 utterances of data from this purely synthetic dataset, we can substantially improve street name transcription. 


\paragraph{Limited Risk of Dual Use} One potential concern is that this technique could be misused to deliberately generate stereotypical or caricatured speech. In practice, however, such misuse is constrained by the behavior of current voice cloning models. The synthesis quality degrades rapidly when the prompt contains large amounts of English text. The model performs best when the prompt is predominantly in the target foreign language with only occasional English insertions. Attempts to generate full English sentences in the style of foreign-language speech tend to collapse into unintelligible audio, limiting the feasibility of producing realistic or scalable imitations of stereotypical speech patterns.

\begin{figure}[]
    \centering
    \includegraphics[width=.95\linewidth]{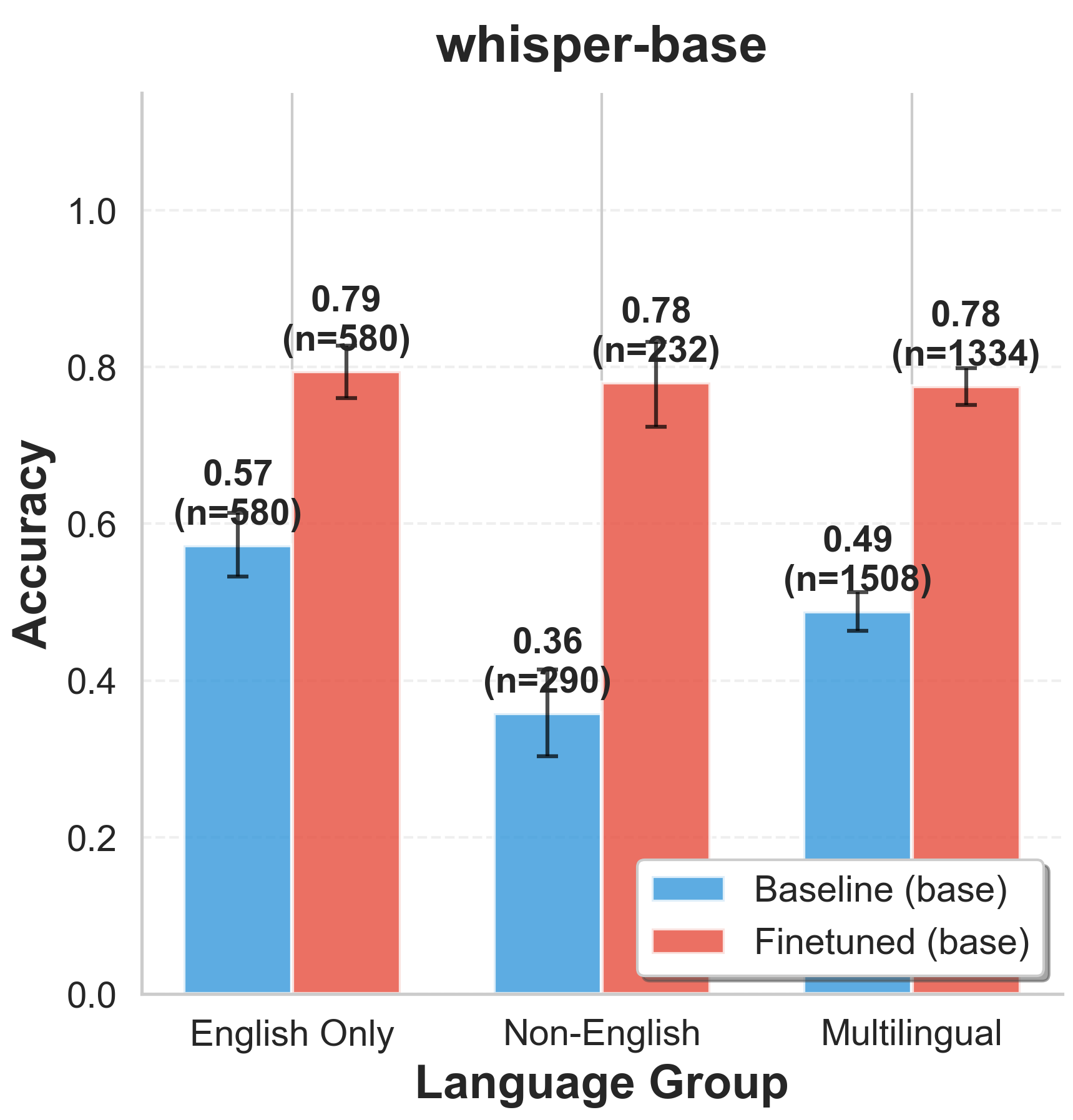}
    \caption{Improvement in accuracy from the finetuned model across language groups. 95\% confidence intervals calculated via bootstrap resampling of 10,000 samples. This holds true for Whisper models across all sizes, Figure \ref{fig:all_whisper_finetuned}}
    \label{fig:finetune_results}
\end{figure}


\paragraph{Out-Of-Distribution Voices}
Despite training on synthetic data, the model improves on real voices from the SF Street datasets, with 116\% and 60\% relative gain from the base model for non-English primary and multilingual speakers, respectively, Figure \ref{fig:finetune_results}. 

\paragraph{Out-Of-Distribution Languages}
Additionally, we see that even for languages that were not originally in the synthetic training data (e.g., languages spoken by participants of SF Streets but not supported by XTTS like Vietnamese), we still see gains in a model's ability to recognize speech from a Vietnamese speaker --- suggesting a generalizability effect from learning other ways of speaking, Figure \ref{fig:finetuned_ooo_language}. 

We additionally train 16 \texttt{Whisper-base} models, each with synthetic data from one language (e.g., training a model with synthetic data from French only) and evaluate how well each of these language-specific models performs on different speakers and street names Figures \ref{fig:finetuned_x_primary_language}, \ref{fig:finetuned_x_streetname}. We find that training on certain languages (e.g., Russian, Arabic, and German) led to higher gains than training on other languages, but find no evidence that training one synthetic language will directly help transcription from speakers of the language. We do see that aggregated training performs best; training the languages individually leads to improvements of 12 - 18\% on average, meaning training with all 16 languages leads to an overall 28\% gain in accuracy.

\begin{figure}[h!]
    \centering
    \includegraphics[width=0.95\linewidth]{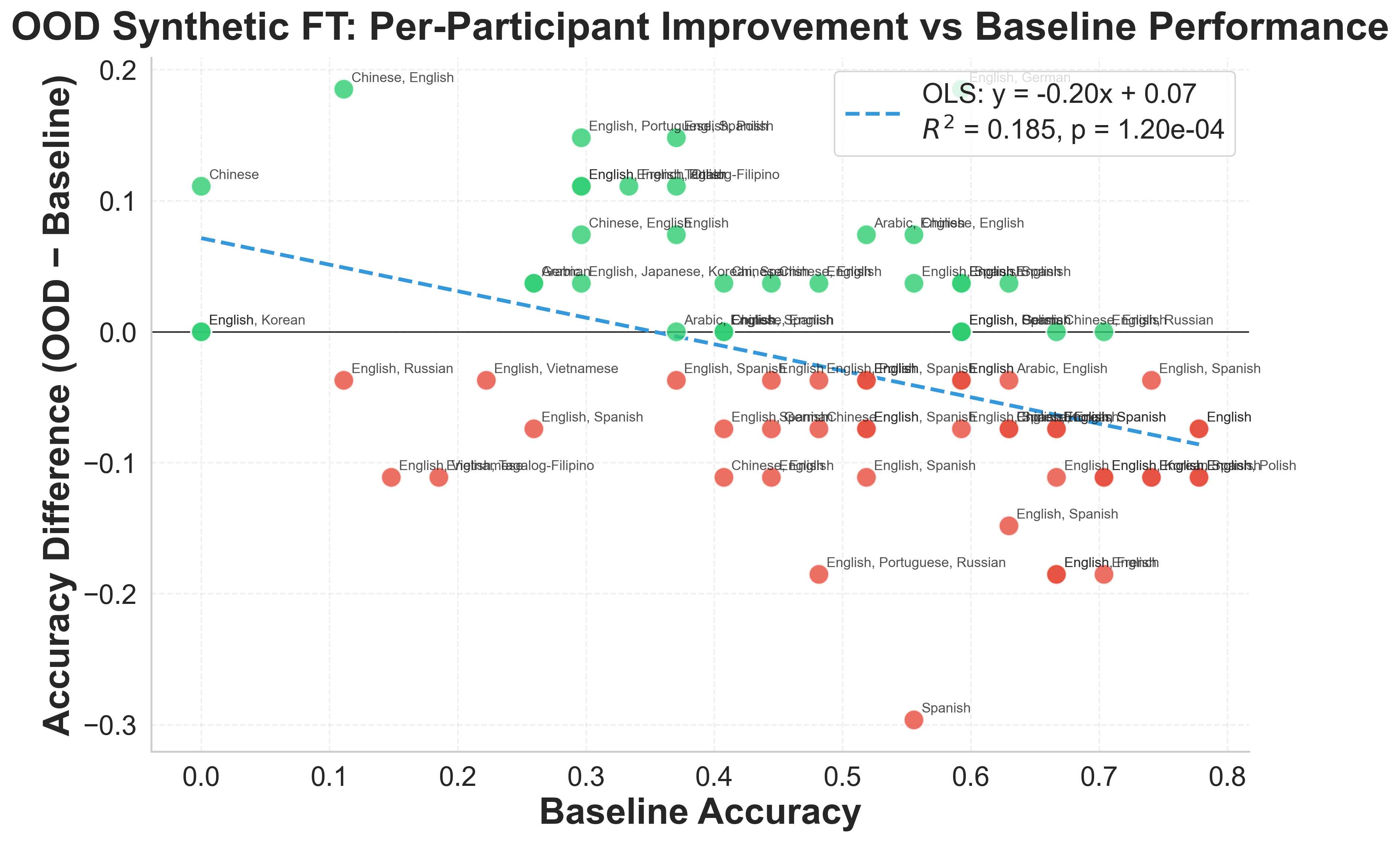}
    \caption{Training only on synthetic out-of-distribution street names}
    \label{fig:ooo_synthetic_improvement}
\end{figure}

\paragraph{Out-Of-Distribution Street Names}
Our data generation process relies entirely on synthetic inputs, and in many cases, complete street name lists can be downloaded directly from municipal websites. As a result, it's lightweight and practical for practitioners to fine-tune language models on every street name within a city to improve transcription rates. However, for the sake of experimentation, we wanted to determine whether it was possible to train on synthetic out-of-distribution street names and observe performance gains on real street name pronunciations. This is a difficult generalization problem because street names are often rare in training data, and their pronunciations can be highly specific to individual entities, limiting transferability.

We finetuned \texttt{Whisper-base} with less than 1,000 synthetic samples and found that overall performance remains largely unchanged ($0.4+ \rightarrow 0.46$). However, as shown in Figure \ref{fig:ooo_synthetic_improvement}, we observed meaningful gains for participants' transcriptions that were poorly transcribed with the baseline model.  The worse the original transcriptions were for the original user, the larger the improvement after finetuning. A linear fit shows that the baseline accuracy explains $R^2 = 0.185$ for the variance in performance change between the fine-tuned and base model, $p< 0.001$. Although these results do not demonstrate broad generalization from synthetic OOD data, they suggest that synthetic training can still substantially improve transcription performance for users—particularly non-native English speakers --- whose voices suffered the worst error rates with baseline models.

\section{Discussion and Conclusion}
In this work, we introduce a new real-world benchmark for evaluating speech recognition systems, focused on the transcription of U.S. street names. Accurate recognition of street names is critical for diverting resources to individuals, and speech models are increasingly deployed in place of human agents for these tasks. However, our evaluations on these models lag behind. This work shows that current evaluation practices fail to adequately capture real-world performance, and our findings demonstrate that even state-of-the-art speech recognition models struggle to correctly transcribe street names. These transcription error rates are furthermore exacerbated in non-English only speakers. To address this gap, we propose a mitigation strategy that leverages publicly available datasets and language models to synthetically generate diverse pronunciations of street names. Using this synthetic data, we fine-tune language models and achieve meaningful improvements in transcription accuracy. We release our datasets as public artifacts to enable benchmarking and further research by the community.

\section*{Acknowledgements}
Thank you to all our online crowdworkers who have contributed to our project! Many thanks to Shang Zhu, Yongchan Kwon, Dan Fu, Sanjana Srivastava, Anna Pot, and Dan Jurafsky for their helpful feedback and support! 

\section*{Impact Statement}
Our work here aims to advance our understanding of language technologies in context. We evaluated several publicly deployed language models and assessed the potential failure modes they present, especially across various demographic groups. We introduce a recipe to synthetically generate named entities with varied pronunciation, leading to substantial gains on this task. We use public datasets and adhere to their terms and agreements, and synthetic voice generation is done via a local, open-sourced model. Lastly, we also aim to release two anonymized public datasets of U.S. speakers pronouncing street names as an artifact and training material for the community.

\bibliography{custom}

@misc{cisa2025ai,
  author       = {{CISA}},
  title        = {Artificial Intelligence in Emergency Communications Centers},
  year         = {2025},
  howpublished = {\url{https://www.cisa.gov/sites/default/files/2025-03/25_0328_s-n_ai-implemen-ecc_infographic_508C.pdf}},
  note         = {Accessed: 2025-12-08}
}

@misc{twilio2025curb,
  author       = {{Twilio}},
  title        = {The future of mobility: how Curb delivers the promised ride with help from Twilio},
  year         = {2025},
  howpublished = {\url{https://customers.twilio.com/en-us/curb}},
  note         = {Accessed: 2025-12-08}
}

@online{carlisle_sf_streets,
  author       = {Carlisle, Henry C.},
  title        = {Early {San Francisco} Street Names - 1846-1849},
  url          = {https://sfmuseum.org/street/stnames2.html},
  urldate      = {2025-12-19}
}

@misc{us_census,
  author       = {{U.S. Census Bureau}},
  title        = {},
  howpublished = {\url{https://data.census.gov}},
  year         = {2022},
  note         = {Accessed 19 Dec. 2025}
}

@misc{sfgov_language_diversity,
  author       = {{City and County of San Francisco}},
  title        = {San Francisco Language Diversity Data},
  howpublished = {\url{https://www.sf.gov/data--san-francisco-language-diversity-data}},
  year         = {2024},
  note         = {Accessed 19 Dec. 2025}
}

@misc{radford2022robustspeechrecognitionlargescale,
      title={Robust Speech Recognition via Large-Scale Weak Supervision}, 
      author={Alec Radford and Jong Wook Kim and Tao Xu and Greg Brockman and Christine McLeavey and Ilya Sutskever},
      year={2022},
      eprint={2212.04356},
      archivePrefix={arXiv},
      primaryClass={eess.AS},
      url={https://arxiv.org/abs/2212.04356}, 
}

@misc{sfmta_taxi_fares,
  author       = {{San Francisco Municipal Transportation Agency}},
  title        = {Taxi Fares},
  howpublished = {\url{https://www.sfmta.com/getting-around/taxi/taxi-fares}},
  year         = {2022},
  note         = {Accessed 18 Dec. 2025}
}

@article{sfexaminer_traffic_2025,
  author       = {Wong, Greg and Wyloge, Evan},
  title        = {{SF} Traffic is Second-Slowest in {US} --- and Getting Worse},
  journal      = {San Francisco Examiner},
  year         = {2025},
  month        = jan,
  day          = {17},
  url          = {https://www.sfexaminer.com/news/transit/sf-traffic-is-second-slowest-in-us-and-getting-worse/article_ebe26770-d45c-11ef-9a49-5fba319c395e.html},
  note         = {Accessed 18 Dec. 2025}
}

@misc{sfmta_taxi_trips,
  author       = {{San Francisco Municipal Transportation Agency}},
  title        = {Average Weekday Taxi Trips},
  howpublished = {\url{https://www.sfmta.com/reports/average-weekday-taxi-trips}},
  year         = {2025},
  note         = { Accessed 18 Dec. 2025}
}

@misc{sfmta_etc,
  author       = {{San Francisco Municipal Transportation Agency}},
  title        = {Essential Trip Card},
  howpublished = {\url{https://www.sfmta.com/getting-around/accessibility/paratransit/essential-trip-card}},
  year         = {2024},
  note         = {Accessed 18 Dec. 2025}
}

@misc{xtts,
      title={XTTS: a Massively Multilingual Zero-Shot Text-to-Speech Model}, 
      author={Edresson Casanova and Kelly Davis and Eren Gölge and Görkem Göknar and Iulian Gulea and Logan Hart and Aya Aljafari and Joshua Meyer and Reuben Morais and Samuel Olayemi and Julian Weber},
      year={2024},
      eprint={2406.04904},
      archivePrefix={arXiv},
      primaryClass={eess.AS},
      url={https://arxiv.org/abs/2406.04904}, 
}

@inproceedings{ardila2020common,
  title={Common voice: A massively-multilingual speech corpus},
  author={Ardila, Rosana and Branson, Megan and Davis, Kelly and Kohler, Michael and Meyer, Josh and Henretty, Michael and Morais, Reuben and Saunders, Lindsay and Tyers, Francis and Weber, Gregor},
  booktitle={Proceedings of the twelfth language resources and evaluation conference},
  pages={4218--4222},
  year={2020}
}

@misc{twilio_gather,
  author       = {{Twilio Inc.}},
  title        = {{TwiML Voice: Gather}},
  year         = {2025},
  url          = {https://www.twilio.com/docs/voice/twiml/gather},
  urldate      = {2026-01-05},
}

@inproceedings{panayotov2015librispeech,
  title={Librispeech: an asr corpus based on public domain audio books},
  author={Panayotov, Vassil and Chen, Guoguo and Povey, Daniel and Khudanpur, Sanjeev},
  booktitle={2015 IEEE international conference on acoustics, speech and signal processing (ICASSP)},
  pages={5206--5210},
  year={2015},
  organization={IEEE}
}

@article{garofolo1993timit,
  title={TIMIT acoustic-phonetic continuous speech corpus},
  author={Garofolo, John S and Lamel, Lori F and Fisher, William M and Pallett, David S and Dahlgren, Nancy L and Zue, Victor and Fiscus, Jonathan G},
  journal={(No Title)},
  year={1993},
  publisher={Linguistic data consortium}
}

@inproceedings{godfrey1992switchboard,
  title={SWITCHBOARD: Telephone speech corpus for research and development},
  author={Godfrey, John J and Holliman, Edward C and McDaniel, Jane},
  booktitle={Acoustics, speech, and signal processing, ieee international conference on},
  volume={1},
  pages={517--520},
  year={1992},
  organization={IEEE Computer Society}
}

@inproceedings{paul1992design,
  title={The design for the Wall Street Journal-based CSR corpus},
  author={Paul, Douglas B and Baker, Janet},
  booktitle={Speech and Natural Language: Proceedings of a Workshop Held at Harriman, New York, February 23-26, 1992},
  year={1992}
}

@article{canavan1997callhome,
  title={Callhome american english speech},
  author={Canavan, Alexandra and Graff, David and Zipperlen, George},
  journal={Linguistic Data Consortium},
  year={1997}
}

@inproceedings{cieri2004fisher,
  title={The Fisher corpus: A resource for the next generations of speech-to-text.},
  author={Cieri, Christopher and Miller, David and Walker, Kevin},
  booktitle={LREC},
  volume={4},
  pages={69--71},
  year={2004}
}

@inproceedings{garg2020hierarchical,
  title={Hierarchical Multi-Stage Word-to-Grapheme Named Entity Corrector for Automatic Speech Recognition.},
  author={Garg, Abhinav and Gupta, Ashutosh and Gowda, Dhananjaya and Singh, Shatrughan and Kim, Chanwoo},
  booktitle={Interspeech},
  pages={1793--1797},
  year={2020}
}

@article{ghannay2018end,
  title={End-to-end named entity extraction from speech},
  author={Ghannay, Sahar and Caubriere, Antoine and Esteve, Yannick and Laurent, Antoine and Morin, Emmanuel},
  journal={arXiv preprint arXiv:1805.12045},
  year={2018}
}

@inproceedings{ning2024breaking,
  title={Breaking the Boundaries: A Unified Framework for Chinese Named Entity Recognition Across Text and Speech},
  author={Ning, Jinzhong and Sun, Yuanyuan and Xu, Bo and Yang, Zhihao and Luo, Ling and Lin, Hongfei},
  booktitle={Findings of the Association for Computational Linguistics: EMNLP 2024},
  pages={1250--1260},
  year={2024}
}

@inproceedings{caubriere2020we,
  title={Where are we in named entity recognition from speech?},
  author={Caubri{\`e}re, Antoine and Rosset, Sophie and Est{\`e}ve, Yannick and Laurent, Antoine and Morin, Emmanuel},
  booktitle={Proceedings of the Twelfth Language Resources and Evaluation Conference},
  pages={4514--4520},
  year={2020}
}

@article{kaufman2016intelligent,
  title={Intelligent paratransit},
  author={Kaufman, Sarah M and Smith, Ashley and O'Connell, Jenny and Marulli, David},
  year={2016}
}

@article{sikder2019uses,
  title={Who uses ride-hailing services in the United States?},
  author={Sikder, Sujan},
  journal={Transportation research record},
  volume={2673},
  number={12},
  pages={40--54},
  year={2019},
  publisher={SAGE Publications Sage CA: Los Angeles, CA}
}

@misc{jiang2019ridehailing,
  author       = {Jiang, Jingjing},
  title        = {More Americans are using ride-hailing apps},
  year         = {2019},
  month        = jan,
  day          = {4},
  publisher    = {Pew Research Center},
  howpublished = {\url{https://www.pewresearch.org/short-reads/2019/01/04/more-americans-are-using-ride-hailing-apps/}},
  note         = {Accessed: 2026-01-27}
}

@article{koenecke2020racial,
  title={Racial disparities in automated speech recognition},
  author={Koenecke, Allison and Nam, Andrew and Lake, Emily and Nudell, Joe and Quartey, Minnie and Mengesha, Zion and Toups, Connor and Rickford, John R and Jurafsky, Dan and Goel, Sharad},
  journal={Proceedings of the national academy of sciences},
  volume={117},
  number={14},
  pages={7684--7689},
  year={2020},
  publisher={National Academy of Sciences}
}

@article{hofmann2024ai,
  title={AI generates covertly racist decisions about people based on their dialect},
  author={Hofmann, Valentin and Kalluri, Pratyusha Ria and Jurafsky, Dan and King, Sharese},
  journal={Nature},
  volume={633},
  number={8028},
  pages={147--154},
  year={2024},
  publisher={Nature Publishing Group UK London}
}

@techreport{sfmta_taxi_survey,
  title = {Best Practices Studies of Taxi Regulation: Taxi User Surveys},
  author = {{San Francisco Municipal Transportation Agency}},
  institution = {San Francisco Municipal Transportation Agency},
  year = {2013},
  url = {https://www.sfmta.com/sites/default/files/Draft%20SF%20UserSurvey%2055%20WEB%20version04042013.pdf}
}

@misc{SF_LanguageDiversityData_SFgov,
  title        = {San Francisco Language Diversity Data},
  author       = {{City and County of San Francisco}},
  year         = {2026},
  howpublished = {\url{https://www.sf.gov/data--san-francisco-language-diversity-data}},
  note         = {Accessed: 2026-02-11},
}
\bibliographystyle{icml2026}

\newpage
\appendix
\section{Appendix}

\label{sec:appendix}

\subsection{SF Streets Public Dataset}
\label{sec:sf_streets_public}
The publicly released SF Streets dataset differs slightly from the dataset analyzed in this paper to respect participants who did not want their data released. It contains 47 of the original participants' data and 45 of newly collected data. The demographic breakdown of the participants in SF Streets public are seen below. 

\begin{table}[h!]
\centering
\small
\renewcommand{\arraystretch}{1.1}
\begin{tabular}{@{}llrr@{}}
\toprule
\textbf{SF Streets} &  & $n$ & \% \\
\midrule
\textit{Gender} & Male & 46 & 50 \\
 & Female & 45 & 48 \\
 & Prefer not to say & 1 & 1 \\
\addlinespace
\textit{Age} & 18--19 & 1 & 1 \\
\textit{(M=37.0, SD=10.7)} & 20--29 & 25 & 27 \\
 & 30--39 & 31 & 33 \\
 & 40--49 & 21 & 22 \\
 & 50--59 & 11 & 11 \\
 & 60--69 & 2 & 2 \\
 & 70+ & 1 & 1 \\
\addlinespace
\textit{Ethnicity} & White & 45 & 48 \\
 & Asian & 31 & 33 \\
 & Other & 8 & 8 \\
 & Mixed & 4 & 4 \\
 & Black & 3 & 3 \\
 & Prefer not to say & 1 & 1 \\
\addlinespace
\textit{Primary Language} & \textcolor[HTML]{599ebc}{English Only} & 12 & 13 \\
 & \textcolor[HTML]{e68600}{Multilingual w/ English} & 66 & 71 \\
 & \textcolor[HTML]{b4638d}{Non-English} & 14 & 15 \\
\bottomrule
\end{tabular}
\caption{Participant demographics for SF streets dataset ($n=92$). The participants' primary languages represented 21 unique languages (Afrikaans, Arabic, Belarusian, Bengali, Cantonese, Chinese, Dutch, English, French, German, Hindi, Korean, Mandarin, Polish, Portuguese, Russian, Spanish, Tagalog-Filipino, Ukrainian, Urdu, Vietnamese).}
\label{tab:demographics}
\end{table}

\begin{figure*}
    \includegraphics[width=\textwidth]{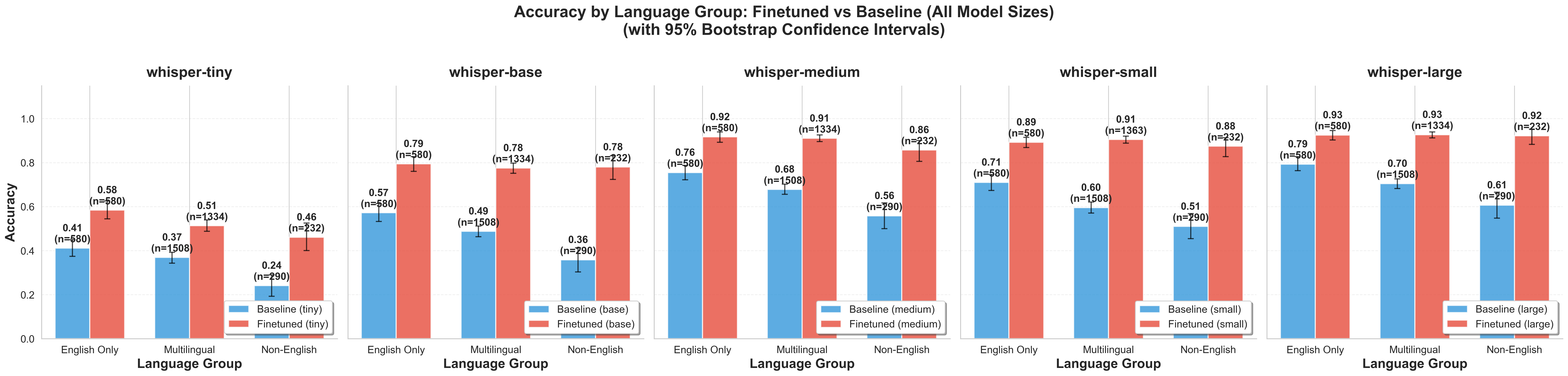}
    \caption{Accuracy between finetuned and baseline models across model sizes}
    \label{fig:all_whisper_finetuned}
\end{figure*}

\begin{figure*}
    \includegraphics[width=\textwidth]
    {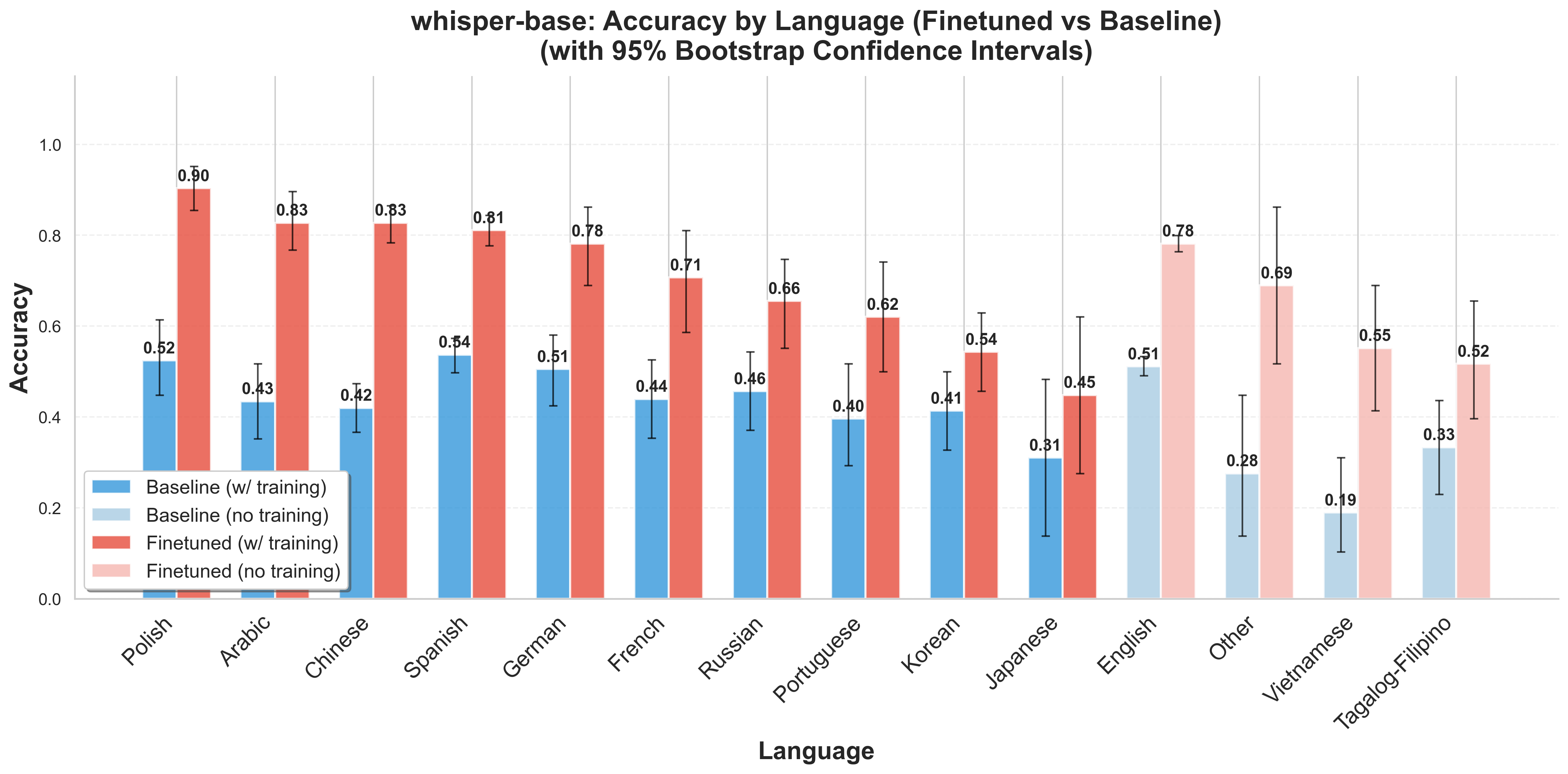}
    \caption{Accuracy between finetuned and baseline models across model sizes}
    \label{fig:finetuned_ooo_language}
\end{figure*}

\begin{figure*}
    \includegraphics[width=\textwidth]
    {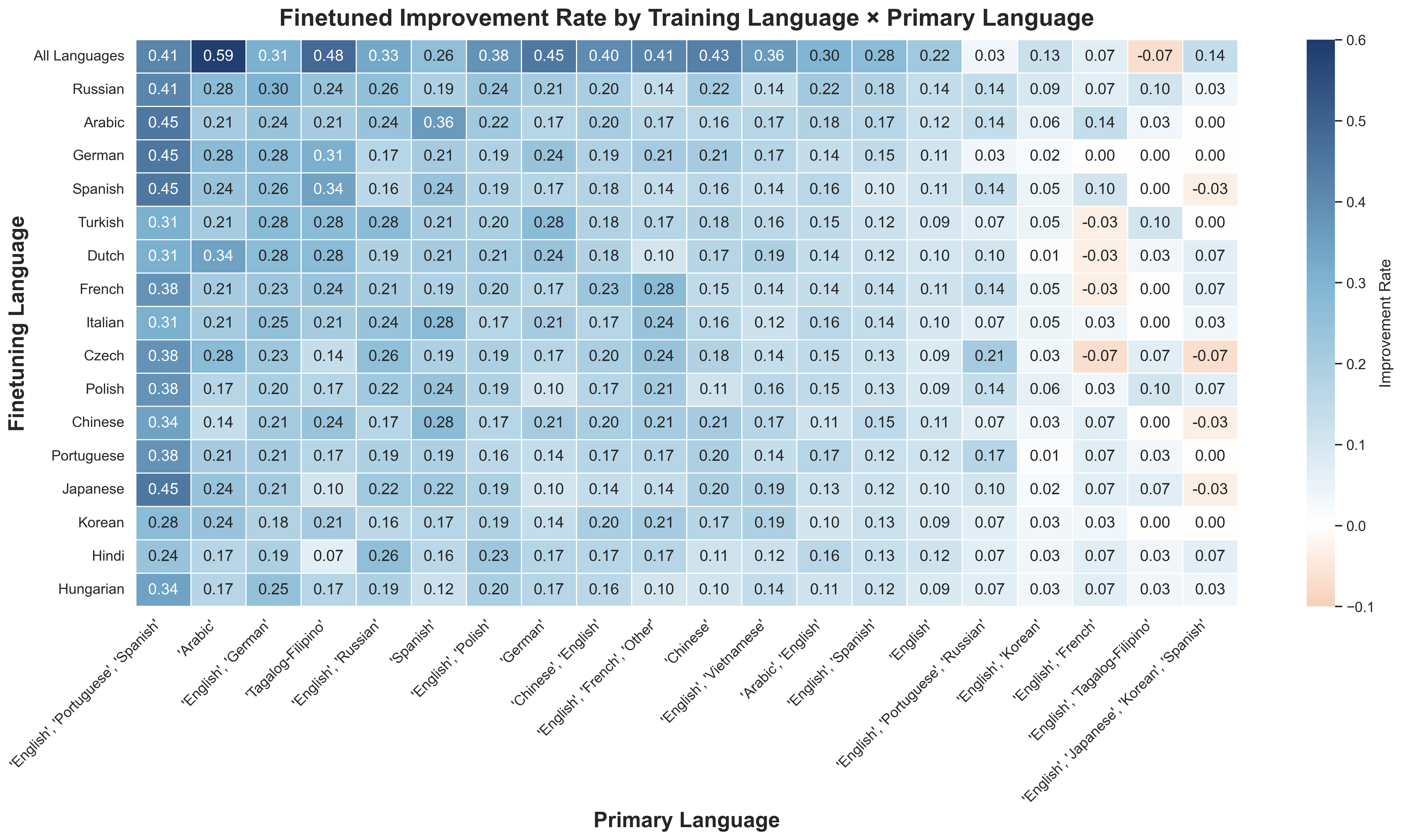}
    \caption{Accuracy between base-whisper and a finetuned model using synthetic data only from one language, crossed by test data of various languages.}
    \label{fig:finetuned_x_primary_language}
\end{figure*}

\begin{figure*}
    \includegraphics[width=\textwidth]
    {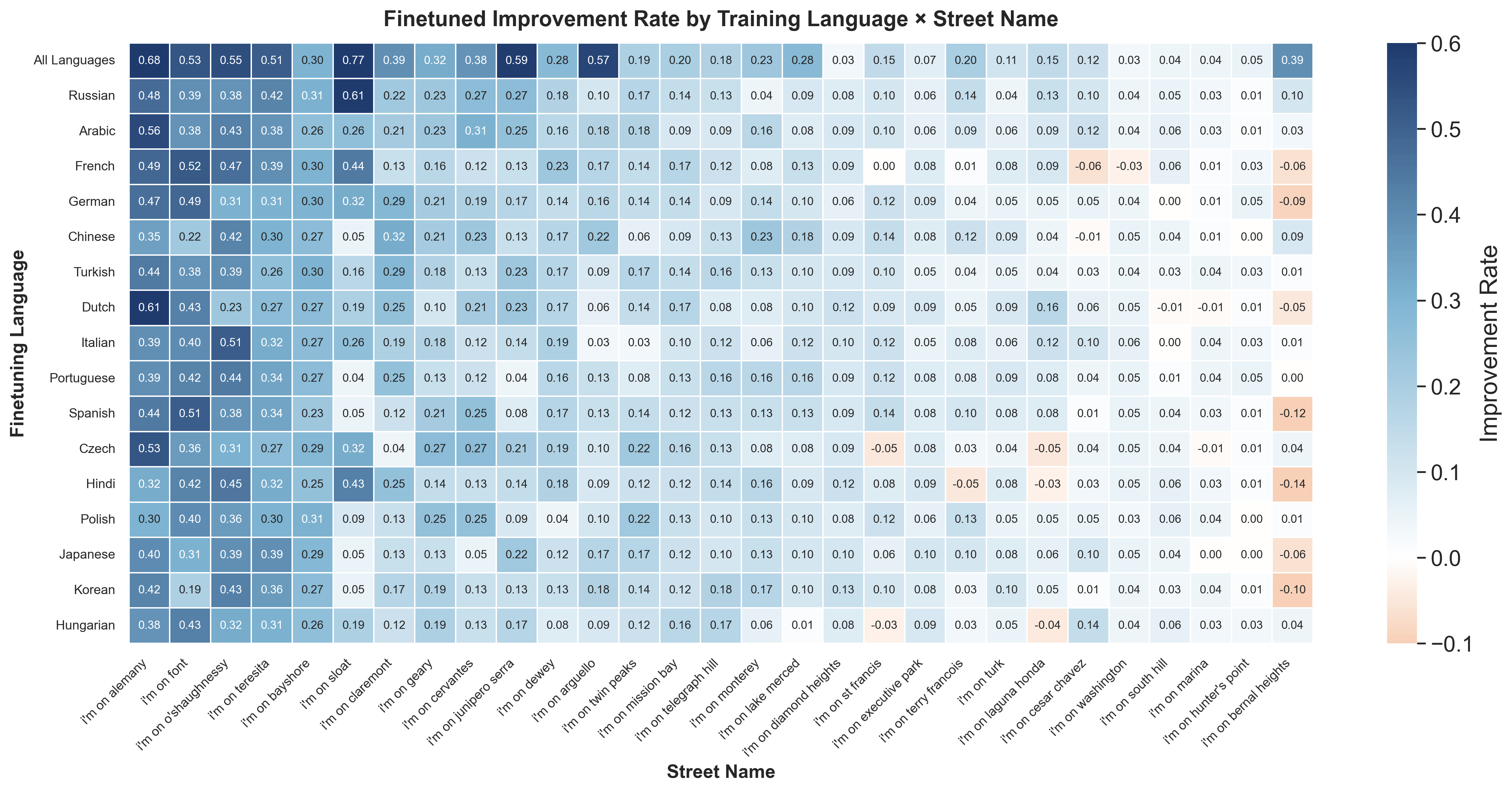}
    \caption{Accuracy between base-whisper and a finetuned model using synthetic data only from one language crossed by names of streets}
    \label{fig:finetuned_x_streetname}
\end{figure*}

\begin{table}[h!]
\centering
\small
\begin{tabular}{ll}
\toprule
\textbf{Street Name} & \textbf{Prompt Format} \\
\midrule
ALEMANY & ``I'm on ALEMANY'' \\
ARGUELLO & ``I'm on ARGUELLO'' \\
BAY SHORE & ``I'm on BAY SHORE'' \\
BERNAL HEIGHTS & ``I'm on BERNAL HEIGHTS'' \\
CERVANTES & ``I'm on CERVANTES'' \\
CESAR CHAVEZ & ``I'm on CESAR CHAVEZ'' \\
CLAREMONT & ``I'm on CLAREMONT'' \\
DEWEY & ``I'm on DEWEY'' \\
DIAMOND HEIGHTS & ``I'm on DIAMOND HEIGHTS'' \\
EXECUTIVE PARK & ``I'm on EXECUTIVE PARK'' \\
FONT & ``I'm on FONT'' \\
GEARY & ``I'm on GEARY'' \\
HUNTERS POINT & ``I'm on HUNTERS POINT'' \\
JUNIPERO SERRA & ``I'm on JUNIPERO SERRA'' \\
LAGUNA HONDA & ``I'm on LAGUNA HONDA'' \\
LAKE MERCED & ``I'm on LAKE MERCED'' \\
MARINA & ``I'm on MARINA'' \\
MISSION BAY & ``I'm on MISSION BAY'' \\
MONTEREY & ``I'm on MONTEREY'' \\
O'SHAUGHNESSY & ``I'm on O'SHAUGHNESSY'' \\
SLOAT & ``I'm on SLOAT'' \\
SOUTH HILL & ``I'm on SOUTH HILL'' \\
ST FRANCIS & ``I'm on ST FRANCIS'' \\
TELEGRAPH HILL & ``I'm on TELEGRAPH HILL'' \\
TERESITA & ``I'm on TERESITA'' \\
TERRY FRANCOIS & ``I'm on TERRY FRANCOIS'' \\
TURK & ``I'm on TURK'' \\
TWIN PEAKS & ``I'm on TWIN PEAKS'' \\
WASHINGTON & ``I'm on WASHINGTON'' \\
\bottomrule
\end{tabular}
\caption{San Francisco street names used in the study. Suffixes are commonly omitted in local signage and postal addresses, and also excluded in our dataset (Table \ref{tab:streetnames}). Instructions to the user was: \textit{Recording yourself saying the following text ONLY ONCE: "I'm on ALEMANY"}}
\label{tab:streetnames}
\end{table}


\begin{table}[h]
\centering
\begin{tabular}{ll}
\toprule
\textbf{Street Name} & \textbf{Accepted Alternative} \\
\midrule
BAYSHORE & BAY\textbf{\_}SHORE \\
HUNTER'S POINT & HUNTE\textbf{RS} POINT \\
CESAR CHAVEZ & C\textbf{AE}SAR CHAVEZ \\
MONTEREY & MONTER\textbf{R}EY \\
CERVANTES & \textbf{S}ERVANTES \\
ALEMANY & A\textbf{LM}ANY \\
ALEMANY & AL\textbf{A}MANY \\
TWIN PEAKS & TWI\textbf{NP}EAKS \\
ARGUELLO & ARGUE\textbf{L}O \\
BERNAL HEIGHTS & B\textbf{U}RNAL HEIGHTS \\
TERRY FRANCOIS & TERRY\textbf{,} FRANCOIS \\
\bottomrule
\end{tabular}
\caption{Phonetically equivalent spellings of street names for the SF Streets Dataset}
\label{tab:SF_streets_alternatives}
\end{table}

\begin{table}[]
\centering
\begin{tabular}{l}
\toprule
\textbf{US Street Prefix Paraphrases} \\
\midrule 
Can you pick me up at [STREET NAME]? \\
Can you come get me on [STREET NAME]? \\
Can you meet me on [STREET NAME]? \\ 
Would you mind picking me up at [STREET NAME]? \\
Could you come get me at [STREET NAME]? \\
Are you able to pick me up at [STREET NAME]? \\
I’m at [STREET NAME] \\
I’m near [STREET NAME] \\
I’m close to [STREET NAME] \\
I’m right by [STREET NAME] \\
I’m in the vicinity of [STREET NAME] \\
I’m currently near [STREET NAME] \\
I live near [STREET NAME] \\
My place is near [STREET NAME] \\
I’m staying near [STREET NAME] \\
I’m based near [STREET NAME] \\
My residence is near [STREET NAME] \\
My place is located on [STREET NAME] \\
\bottomrule
\end{tabular}
\caption{Prefix paraphrases for the US streets dataset.}
\label{tab:paraphrases}
\end{table}
\begin{table*}[]
\tiny
\centering

\begin{tabular}{llll}
\toprule
phrase & city & phrase & city \\
\midrule
Are you able to pick me up at STEVENSON? & Chicago & Can you pick me up at STRATO? & Jacksonville \\
I’m at THOMAS. & Chicago & Are you able to pick me up at FABRAY? & Jacksonville \\
I’m currently near WAVELAND. & Chicago & I’m currently near PLANTS. & Jacksonville \\
I’m at MIDWAY AIRPORT LOWER. & Chicago & Are you able to pick me up at RUBY? & Jacksonville \\
I’m close to FRONTIER. & Chicago & I’m near KINGS COLONY. & Jacksonville \\
Would you mind picking me up at SOLIDARITY? & Chicago & I’m close to COXWELL. & Jacksonville \\
I’m close to LELAND. & Chicago & I’m based near TAMRA. & Jacksonville \\
I’m right by LAWNDALE. & Chicago & My place is located on MANNING. & Jacksonville \\
Would you mind picking me up at DAN RYAN CANALPORT? & Chicago & Would you mind picking me up at LADI? & Jacksonville \\
I’m close to BANKS. & Chicago & I live near WAGES WAY. & Jacksonville \\
I’m in the vicinity of PARKER. & Chicago & I’m based near BLUE RIDGE. & Jacksonville \\
I’m staying near REDFIELD. & Chicago & Could you come get me at FENNELL? & Jacksonville \\
Can you pick me up at CATALPA? & Chicago & I’m in the vicinity of TOWHEE. & Jacksonville \\
Would you mind picking me up at LEHMANN? & Chicago & Can you meet me on DARROW? & Jacksonville \\
I’m close to KENNEDY JACKSON BL. & Chicago & I’m at MATTHEW. & Jacksonville \\
I’m currently near GIDDINGS. & Chicago & Can you pick me up at HOMEPORT? & Jacksonville \\
Would you mind picking me up at GRENSHAW? & Chicago & My place is near SEMINARY. & Jacksonville \\
I’m currently near WALLER. & Chicago & I’m in the vicinity of OAK LAWN. & Jacksonville \\
I’m based near MAGNET. & Chicago & I’m right by DEEPWOOD. & Jacksonville \\
My place is located on BURLING. & Chicago & Can you pick me up at GUN CLUB? & Jacksonville \\
Could you come get me at BRIAR? & Chicago & Can you meet me on BOYSENBERRY? & Jacksonville \\
I’m staying near DEAN. & Chicago & I’m close to ALVIN ROAD. & Jacksonville \\
My place is located on DAN RYAN. & Chicago & I’m currently near KERNAN MILL. & Jacksonville \\
Can you come get me on TRUMBULL? & Chicago & Can you meet me on DEEPWOOD? & Jacksonville \\
I’m near PRINDIVILLE. & Chicago & I’m currently near SIMMS COVE. & Jacksonville \\
My place is located on OLD LAKE. & Chicago & Can you meet me on HUNTERS LAKE? & Jacksonville \\
I’m currently near LSD FORT DEARBORN. & Chicago & Are you able to pick me up at DARTMOUTH? & Jacksonville \\
My place is located on DAWSON. & Chicago & My place is near DAHLONEGA. & Jacksonville \\
My residence is near NATCHEZ. & Chicago & I’m right by CORNELL. & Jacksonville \\
Would you mind picking me up at BALBO? & Chicago & My place is near HOLLOW PINE. & Jacksonville \\
I’m right by WOODACRE. & Dallas & I live near REDWOOD. & Los Angeles \\
I’m staying near GAYGLEN. & Dallas & I’m at ELECTRIC. & Los Angeles \\
Would you mind picking me up at BRIARCREST? & Dallas & My place is near MALONE. & Los Angeles \\
My place is near LAKEFAIR. & Dallas & I’m staying near BRUCE. & Los Angeles \\
Can you pick me up at MENEFEE? & Dallas & I’m based near FALMOUTH. & Los Angeles \\
Would you mind picking me up at HAMBLEN? & Dallas & Can you pick me up at BARRYWOOD? & Los Angeles \\
I live near ABLES. & Dallas & Could you come get me at HANLEY? & Los Angeles \\
My residence is near MAIL. & Dallas & I’m currently near LEONARD. & Los Angeles \\
Can you come get me on MOSER? & Dallas & My place is near CAMINO REAL. & Los Angeles \\
I’m currently near GLENAIRE. & Dallas & My residence is near OAK HILL. & Los Angeles \\
I’m staying near RENAISSANCE. & Dallas & I’m near GAZETTE. & Los Angeles \\
I live near DUNSTAN. & Dallas & I’m currently near FORTUNA. & Los Angeles \\
Could you come get me at GRIFFIN? & Dallas & My place is near BLADES. & Los Angeles \\
Can you pick me up at CRESTEDGE? & Dallas & I’m near AMAR. & Los Angeles \\
I’m near REUNION. & Dallas & Can you pick me up at TERI? & Los Angeles \\
I’m close to OXHILL. & Dallas & Could you come get me at MADELIA? & Los Angeles \\
I’m close to MILLBROOK. & Dallas & Are you able to pick me up at BELCANTO? & Los Angeles \\
I live near CROWNOVER. & Dallas & I’m currently near STONE OAK. & Los Angeles \\
I’m close to VISTADALE. & Dallas & My residence is near GARVANZA. & Los Angeles \\
I’m based near SABINE. & Dallas & Would you mind picking me up at HUNT CLUB? & Los Angeles \\
My place is located on SYLVIA. & Dallas & Can you come get me on ARGYLE? & Los Angeles \\
Can you come get me on WOODBEND? & Dallas & I live near LEXICON. & Los Angeles \\
I’m in the vicinity of OAK HILL. & Dallas & I’m staying near ZORADA. & Los Angeles \\
I’m staying near NEMECHEK. & Dallas & My residence is near MENDIPS RIDGE. & Los Angeles \\
Would you mind picking me up at OAKLAND CEMETERY? & Dallas & My place is located on MARLBORO. & Los Angeles \\
Can you pick me up at BLACK WALNUT? & Dallas & Can you come get me on VESTONE? & Los Angeles \\
I’m near LODGE. & Dallas & Would you mind picking me up at COLLIER? & Los Angeles \\
I live near CONCORDANT. & Dallas & I’m near OCAMPO. & Los Angeles \\
I’m staying near KELLY. & Dallas & I’m at LAURA. & Los Angeles \\
My place is located on DUNHILL. & Dallas & I’m currently near LAS FLORES. & Los Angeles \\
Would you mind picking me up at FINCH FALLS? & Houston & Can you come get me on LOCKWOOD? & New York City \\
I’m in the vicinity of NEW ENGLAND. & Houston & Can you meet me on MOORE? & New York City \\
My place is near MACROOM. & Houston & I live near MARBLE CEMETERY BOUNDARY. & New York City \\
My place is located on NATURES SONG. & Houston & Can you come get me on BROWN? & New York City \\
Can you come get me on DARBYDALE? & Houston & I’m near ECHO. & New York City \\
Could you come get me at GRANITE ROCK? & Houston & I’m staying near PATH OF PROGRESS. & New York City \\
I’m in the vicinity of HEATHER HILL. & Houston & My residence is near INEZ. & New York City \\
I’m right by MESONES. & Houston & My place is located on ERBEN. & New York City \\
My residence is near WESTBRAE OAKS. & Houston & Are you able to pick me up at CHURCH? & New York City \\
I’m staying near BARON BROOK. & Houston & Are you able to pick me up at POINT? & New York City \\
I live near BOATBILL. & Houston & I’m near BARRETT PARK BOUNDARY. & New York City \\
Can you come get me on BLUEBIRD? & Houston & My place is near DOANE. & New York City \\
Can you come get me on MAGNOLIA POINT? & Houston & Can you meet me on PINEAPPLE? & New York City \\
Can you meet me on CLAIRY? & Houston & Could you come get me at MEADOW LAKE PROMENADE? & New York City \\
My place is near BRIXHAM OAKS. & Houston & I’m based near PERCIVAL. & New York City \\
I’m near ESSENDINE. & Houston & My place is located on MAYVILLE. & New York City \\
My residence is near CRESCENT PASS. & Houston & Would you mind picking me up at QUENTIN? & New York City \\
Could you come get me at MIRANDOLA? & Houston & I’m at FLUSHING BAY SHORELINE. & New York City \\
My place is near WENSLEY. & Houston & I’m right by MARION. & New York City \\
My residence is near TROTTING. & Houston & I’m currently near BARDWELL. & New York City \\
Are you able to pick me up at AUTUMN LEAF? & Houston & I’m near REMINGTON. & New York City \\
I’m near PRELUDE. & Houston & I’m right by CHAUNCEY. & New York City \\
I’m in the vicinity of DANFORTH. & Houston & I live near BURNETT. & New York City \\
Can you pick me up at BECKER CEMETERY? & Houston & I’m right by WIELAND. & New York City \\
I’m close to SUNDANCE SPRINGS. & Houston & I’m currently near CONCORD. & New York City \\
I’m in the vicinity of ROYAL BAY. & Houston & I’m staying near FLATLANDS. & New York City \\
I’m based near CADENA. & Houston & Can you come get me on VANDERVOORT? & New York City \\
Can you come get me on QUIET? & Houston & I’m right by CARLY. & New York City \\
I’m in the vicinity of HARPER. & Houston & My place is near ERICSSON. & New York City \\
Are you able to pick me up at LYRIC? & Houston & Can you pick me up at ARCHER? & New York City \\
\bottomrule
\end{tabular}
\caption{Dataset for U.S. Streets}
\label{tab:US_streets_dataset}
\end{table*}

\begin{table*}[]
\tiny
\centering

\begin{tabular}{llll}
\toprule
phrase & city & phrase & city \\
\midrule
Can you meet me on BROWNS? & Philadelphia & I’m staying near MANOCK. & San Diego \\
My place is near NICE. & Philadelphia & Can you pick me up at SANTOLINA? & San Diego \\
Can you meet me on NORTHWOOD? & Philadelphia & I’m at BROOKLYN. & San Diego \\
I’m currently near NICETOWN. & Philadelphia & I’m in the vicinity of BERYL COVE. & San Diego \\
I’m in the vicinity of WORRELL. & Philadelphia & Can you meet me on WOMBLE? & San Diego \\
I live near COWDEN. & Philadelphia & My residence is near CARIB. & San Diego \\
I’m based near FLAGLER. & Philadelphia & Can you come get me on STEEL? & San Diego \\
My place is located on HAYDEN. & Philadelphia & Could you come get me at LOMICA? & San Diego \\
My place is near BONAPARTE. & Philadelphia & I’m at ARROYO SECO. & San Diego \\
Are you able to pick me up at FROST? & Philadelphia & I’m right by AMBER LAKE. & San Diego \\
I’m right by SHAW. & Philadelphia & I’m staying near DALEWOOD. & San Diego \\
I’m in the vicinity of GODFREY. & Philadelphia & Are you able to pick me up at LA TRUCHA? & San Diego \\
Are you able to pick me up at BEULAH? & Philadelphia & I live near DERBY. & San Diego \\
I’m close to MICA. & Philadelphia & My place is near VINTAGE. & San Diego \\
Can you pick me up at HARVARD? & Philadelphia & I’m staying near ENTERPRISE. & San Diego \\
My residence is near HAYNADIER. & Philadelphia & Can you come get me on EDEN? & San Diego \\
Can you come get me on POTTERTON? & Philadelphia & I’m at CATES. & San Diego \\
I’m in the vicinity of MONUMENT. & Philadelphia & I’m in the vicinity of TOWN VIEW. & San Diego \\
Are you able to pick me up at GREENHILL APARTMENT? & Philadelphia & I’m currently near EARL. & San Diego \\
I’m staying near LEE LYNN. & Philadelphia & Could you come get me at MOONSTONE? & San Diego \\
Can you pick me up at LONEY? & Philadelphia & I’m at GERMAINE. & San Diego \\
I’m near ORTHODOX. & Philadelphia & My place is near CANTERBURY. & San Diego \\
Would you mind picking me up at SHIPWAY? & Philadelphia & I’m staying near FONTANELLE. & San Diego \\
I’m near KEYSTONE. & Philadelphia & My place is near CAMINITO BASSWOOD. & San Diego \\
I’m right by RIPLEY. & Philadelphia & Could you come get me at MEDOC? & San Diego \\
Can you come get me on BURNHAM? & Philadelphia & My residence is near PALERO. & San Diego \\
I’m at AMBRIDGE. & Philadelphia & I’m near BERT. & San Diego \\
I’m staying near KRAMS. & Philadelphia & I’m close to TUTHER. & San Diego \\
I’m staying near DREER. & Philadelphia & I’m currently near PARIS. & San Diego \\
Could you come get me at RENO? & Philadelphia & I’m based near MOUNT SAINT HELENS. & San Diego \\
I’m close to VIA DE PEDRO MIGUEL. & Phoenix & Can you pick me up at GLORIA? & San Francisco \\
Could you come get me at RAPALO? & Phoenix & Could you come get me at SLOAN? & San Francisco \\
I’m based near HOPI. & Phoenix & I’m based near MOULTRIE. & San Francisco \\
Could you come get me at DUSTY WREN? & Phoenix & I’m right by EMERY. & San Francisco \\
I’m close to BROADMOOR. & Phoenix & I’m in the vicinity of DARRELL. & San Francisco \\
I’m at WHITE PINE. & Phoenix & My residence is near CLOVER. & San Francisco \\
I’m near SKY TRAIN. & Phoenix & My place is located on OSGOOD. & San Francisco \\
I’m currently near GOLDEN VISTA. & Phoenix & Can you meet me on ELK? & San Francisco \\
Are you able to pick me up at LANDIS? & Phoenix & I’m right by NELSON RISING. & San Francisco \\
I’m near KELTON. & Phoenix & My place is located on SOULE. & San Francisco \\
Can you pick me up at PRAIANO? & Phoenix & Can you pick me up at MCLAREN? & San Francisco \\
Can you meet me on ANNE? & Phoenix & Can you come get me on MAGELLAN? & San Francisco \\
Can you pick me up at BURNING TREE? & Phoenix & Would you mind picking me up at GREENWICH? & San Francisco \\
Are you able to pick me up at FREEMONT? & Phoenix & I’m near BERTHA. & San Francisco \\
I’m based near CHISUM. & Phoenix & Can you come get me on ALLISON? & San Francisco \\
Can you meet me on DAI? & Phoenix & Can you meet me on ELSIE? & San Francisco \\
Are you able to pick me up at KERBY? & Phoenix & I’m at BONNIE BRAE. & San Francisco \\
I’m close to RUNWAY. & Phoenix & I’m staying near RAYCLIFF. & San Francisco \\
My place is located on CULVER. & Phoenix & I’m right by WILLIAMS. & San Francisco \\
I’m in the vicinity of GALLOP. & Phoenix & I’m in the vicinity of BEMIS. & San Francisco \\
I live near CHERYL. & Phoenix & I’m based near NEWMAN. & San Francisco \\
My place is near CAMINO DEL RIO. & Phoenix & My place is near BUCKINGHAM. & San Francisco \\
Could you come get me at CREEDANCE? & Phoenix & Could you come get me at NATICK? & San Francisco \\
I live near CARIBE. & Phoenix & Can you meet me on TURNER? & San Francisco \\
My residence is near MEADOW. & Phoenix & I’m at ANNIE. & San Francisco \\
Would you mind picking me up at MARIPOSA? & Phoenix & My residence is near EMERALD. & San Francisco \\
My place is located on CLINTON. & Phoenix & Can you pick me up at VARENNES? & San Francisco \\
Can you come get me on TURF PARADISE? & Phoenix & Are you able to pick me up at HARRIET? & San Francisco \\
Can you meet me on DANBURY? & Phoenix & I live near RINGOLD. & San Francisco \\
I’m close to GOLDEN PUMA. & Phoenix & I live near JENNINGS. & San Francisco \\
I’m near CRESCENT CHASE. & San Antonio & My residence is near ANNERLY. & San Jose \\
I’m near DREAMWOOD. & San Antonio & I live near CAPELLA. & San Jose \\
I’m currently near BIRNAM OAKS. & San Antonio & I’m based near TWILIGHT. & San Jose \\
Can you come get me on FOREST OAK? & San Antonio & Could you come get me at TONI? & San Jose \\
I’m at MORNING. & San Antonio & Can you meet me on GIBRALTAR? & San Jose \\
I’m right by BISON. & San Antonio & My place is located on IRENE. & San Jose \\
Are you able to pick me up at BASKET ELM? & San Antonio & I’m based near GRAHAM. & San Jose \\
I’m close to CASTLE POND. & San Antonio & I’m in the vicinity of BRIGHTSIDE. & San Jose \\
I live near ETON GREEN. & San Antonio & My residence is near BENZO. & San Jose \\
I’m at KUDU. & San Antonio & My place is located on TOWN. & San Jose \\
I’m right by PIPERS. & San Antonio & Would you mind picking me up at RUE CHENE? & San Jose \\
My place is near BLAZER. & San Antonio & My place is located on HARLISS. & San Jose \\
Can you meet me on VALLEY VIEW? & San Antonio & Would you mind picking me up at KODIAC? & San Jose \\
I’m in the vicinity of PARKVIEW. & San Antonio & I’m staying near ALMOND. & San Jose \\
My residence is near MIRANDA. & San Antonio & I’m right by PIAZZA. & San Jose \\
I’m at ELBEL. & San Antonio & I’m based near BRIONA. & San Jose \\
My residence is near CAMINO VENADO. & San Antonio & My place is located on MAXWELL. & San Jose \\
I live near GREEN ACRES WOODS. & San Antonio & Would you mind picking me up at COURTNEY? & San Jose \\
I’m close to CROW WING. & San Antonio & I’m at BEACON. & San Jose \\
I’m close to WELLOCK. & San Antonio & I’m based near SHERBROOKE. & San Jose \\
Can you meet me on CHESTNUT VIEW? & San Antonio & Can you pick me up at FOREST? & San Jose \\
Would you mind picking me up at DEBBIE? & San Antonio & Are you able to pick me up at STANHOPE? & San Jose \\
Could you come get me at MILL? & San Antonio & I’m currently near BLUE DOLPHIN. & San Jose \\
I’m right by MOSS. & San Antonio & I live near HOMEWARD. & San Jose \\
Are you able to pick me up at ATLAS? & San Antonio & I’m based near CHARLESTON. & San Jose \\
I’m at CONDALIA. & San Antonio & Can you meet me on BARRYMORE? & San Jose \\
I’m based near FALLOW. & San Antonio & Could you come get me at GLEN DONEGAL? & San Jose \\
I’m close to CLAUDIA. & San Antonio & I’m based near DE LA PENA. & San Jose \\
I’m right by BRIDGEPORT. & San Antonio & I’m at SONATA. & San Jose \\
I’m in the vicinity of CASSIANO. & San Antonio & My residence is near VIERRA. & San Jose \\
\bottomrule
\end{tabular}
\caption{Dataset for U.S. Streets}
\label{tab:US_streets_dataset}
\end{table*}

\end{document}